\documentclass[11pt]{article}

\usepackage[]{acl}

\usepackage{times}
\usepackage{latexsym}

\usepackage[T1]{fontenc}
\usepackage[utf8]{inputenc}

\usepackage{microtype}

\usepackage{booktabs}
\usepackage{xcolor}         %
\usepackage{graphicx}
\usepackage{xspace}
\usepackage{todonotes}
\usepackage{multirow}
\usepackage{wrapfig}
\usepackage{caption}
\usepackage{subcaption}
\usepackage{tabularx}
\usepackage{amsmath}
\usepackage{bbm}
\usepackage{bm}
\usepackage{mathtools}
\usepackage{lipsum}
\usepackage[ruled,vlined]{algorithm2e}

\usepackage{pifont}%
\newcommand{\cmark}{\ding{51}}%
\newcommand{\xmark}{\ding{55}}%

\newcommand{\ie}{i.e.,\xspace}
\newcommand{\eg}{e.g.,\xspace}
\newcommand{\eat}[1]{}

\newcommand{\baby}{LaPraDoR\xspace}

\title{\includegraphics[width=0.7cm]{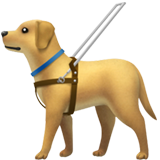} LaPraDoR: Unsupervised Pretrained Dense Retriever \\ for Zero-Shot Text Retrieval }

\author{Canwen Xu$^1$\thanks{\ \ Equal contribution.}\ , Daya Guo$^{2*}$, Nan Duan$^{3}$, Julian McAuley$^1$ \\
$^1$University of California, San Diego, $^2$Sun Yat-sen University, $^3$Microsoft Research Asia\\
$^1$\texttt{\{cxu,jmcauley\}@ucsd.edu}, $^2$\texttt{guody5@mail2.sysu.edu.cn}, \\ $^3$\texttt{nanduan@microsoft.com}\\
}

\begin{document}
\maketitle 
\begin{abstract}
In this paper, we propose LaPraDoR, a pretrained dual-tower dense retriever that does not require any supervised data for training. Specifically, we first present Iterative Contrastive Learning (ICoL) that iteratively trains the query and document encoders with a cache mechanism. 
ICoL not only enlarges the number of negative instances but also keeps representations of cached examples in the same hidden space.
We then propose Lexicon-Enhanced Dense Retrieval (LEDR) as a simple yet effective way to enhance dense retrieval with lexical matching. We evaluate \baby on the recently proposed BEIR benchmark, including 18 datasets of 9 zero-shot text retrieval tasks. Experimental results show that \baby achieves state-of-the-art performance compared with supervised dense retrieval models, and further analysis reveals the effectiveness of our training strategy and objectives. Compared to re-ranking, our lexicon-enhanced approach can be run in milliseconds (22.5$\times$ faster) while achieving superior performance.\footnote{Code and pretrained weights can be found at \url{https://github.com/JetRunner/LaPraDoR}.}
\end{abstract}

\section{Introduction}

Dense retrieval uses dense vectors to represent documents and retrieve documents by similarity scores between query vectors and document vectors. Different from cross-encoders~\citep{sbert,gao2020modularized,macavaney2020efficient} or late-interaction models~\citep{colbert,coil}, which predict a match score for each query-document pair thus are computationally costly, dense retrieval can be run in milliseconds, with the help of an approximate nearest neighbor (ANN) retrieval 
library, e.g., FAISS~\citep{faiss}.

As a drawback, dense retrieval models often require large supervised datasets like MS-MARCO~\citep{msmarco} (533k training examples) or NQ~\citep{47761} (133k training examples) for training. Unfortunately, \citet{thakur2021beir} empirically show that models trained on one dataset suffer from an
out-of-domain (OOD) problem when transferring to another.  This
hinders the applications of dense retrieval systems. On the other hand, creating a large supervised training dataset for dense retrieval is time-consuming and expensive. For many low-resource languages, there is even no existing supervised dataset for retrieval and it can be extremely difficult to construct one.

The recently proposed BEIR benchmark~\citep{thakur2021beir} highlights the 
generalization ability of
text retrieval systems. The benchmark features a setting where models are trained on a large supervised dataset MS-MARCO~\citep{msmarco} and then tested on 18 heterogeneous datasets of 9 tasks. %
In this paper, we propose
\textbf{La}rge-scale \textbf{Pr}etr\textbf{a}ined \textbf{D}ense Zer\textbf{o}-shot \textbf{R}etriever (\baby), a fully unsupervised pretrained retriever for zero-shot text retrieval. While existing dense retrievers need large supervised data and struggle to compete with a lexical matching approach like BM25~\citep{bm25} for zero-shot retrieval, we take a different approach by complementing lexical matching with semantic matching. Without any supervised data, \baby outperforms
all dense retrievers on BEIR. \baby achieves state-of-the-art performance on BEIR with a further fine-tuning, outperforming re-ranking, despite being 22.5$\times$ and 42$\times$ faster on GPU and CPU, respectively.

Training \baby faces two challenges: \textbf{(1) Training Efficiency.} For large-scale pretraining, training efficiency can be important. In contrastive learning, more negative instances often lead to better performance~\citep{declutr,wu2020clear,gao2021scaling}. However, traditional in-batch negative sampling is bottlenecked by limited GPU memory. To alleviate this problem, we propose Iterative Contrastive Learning (ICoL), which iteratively trains the query and document encoders with a cache mechanism. Compared to existing solutions MoCo~\citep{moco} and xMoCo~\citep{xmoco}, ICoL does not introduce extra encoders and can solve the mismatching between representation spaces, thus demonstrating superior performance. \textbf{(2) Versatility.} There are different types of downstream tasks from various domains in both BEIR and real-world applications. We use a large-scale multi-domain corpus, C4~\citep{t5}, to train our \baby model. To make \baby versatile, besides conventional query-document retrieval, we also incorporate document-query, query-query, and document-document retrieval into the pretraining objective. We further share the weights between the query and document encoders and obtain an all-around encoder that fits all retrieval tasks.

To summarize, our contribution is three-fold: (1) We train \baby, an all-around unsupervised pretrained dense retriever that achieves state-of-the-art performance on the BEIR benchmark. (2) We propose Iterative Contrastive Learning (ICoL) for training a retrieval model effectively. (3) We propose Lexicon-Enhanced Dense Retrieval as an efficient way for combining BM25 with a dense retriever, compared to the widely-used re-ranking paradigm.

\section{Related Work}
\paragraph{Dense Retrieval}
DPR~\citep{dpr} initializes a bi-encoder model with BERT~\citep{bert} and achieves better results than earlier dense retrieval methods. RocketQA~\citep{rocketqa} exploits a trained retriever to mine hard negatives and then re-train a retriever with the mined negatives. ANCE~\citep{ance} dynamically mines hard negatives throughout training but requires periodic encoding of the entire corpus. TAS-B~\citep{tasb} is a bi-encoder trained with balanced topic-aware sampling and knowledge distillation from a cross-encoder and a ColBERT model~\citep{colbert}, in addition to in-batch negatives. xMoCo~\citep{xmoco} adapt MoCo~\citep{moco}, a contrastive learning algorithm that is originally proposed for image representation, to text retrieval by doubling its fast and slow encoders. Although these dense retrieval systems demonstrate effectiveness on some datasets, the BEIR benchmark~\citep{thakur2021beir} highlights a main drawback of these dense retrieval systems - failure to generalize to out-of-domain data. This motivates pretraining as a solution for better domain generalization~\citep{gururangan2020dont}. Dense retrieval has also been applied in many other tasks~\citep{day2,day1}.

\paragraph{Pretraining for Retrieval}
\citet{lee2019latent} first propose to pretrain a bi-encoder retriever with an Inverse Cloze Task (ICT), which constructs a training pair by randomly selecting a sentence from a passage as the query and leaving the rest as the document. \citet{ict} propose two pretraining tasks for Wikipedia and attempt to combine them with ICT and masked language modeling (MLM). \citet{guu2020realm} pretrain a retriever and a reader together for end-to-end question answering (QA). Very recently, DPR-PAQ~\citep{ouguz2021domain} highlight the importance of domain matching by using both synthetic and crawled QA data to pretrain and then fine-tune the model on downstream datasets for dialogue retrieval. Condenser~\citep{condenser} is a new Transformer variant for MLM pretraining. It exploits an information bottleneck to facilitate learning for information aggregation. On top of that, coCondenser~\citep{cocondenser} adds an unsupervised corpus-level contrastive loss to warm up the passage embedding space. Different from these works, \baby is the first pretrained retriever that %
does not require fine-tuning on a downstream dataset and can perform zero-shot retrieval.

\begin{figure}
    \centering
    \includegraphics[width=0.9\linewidth]{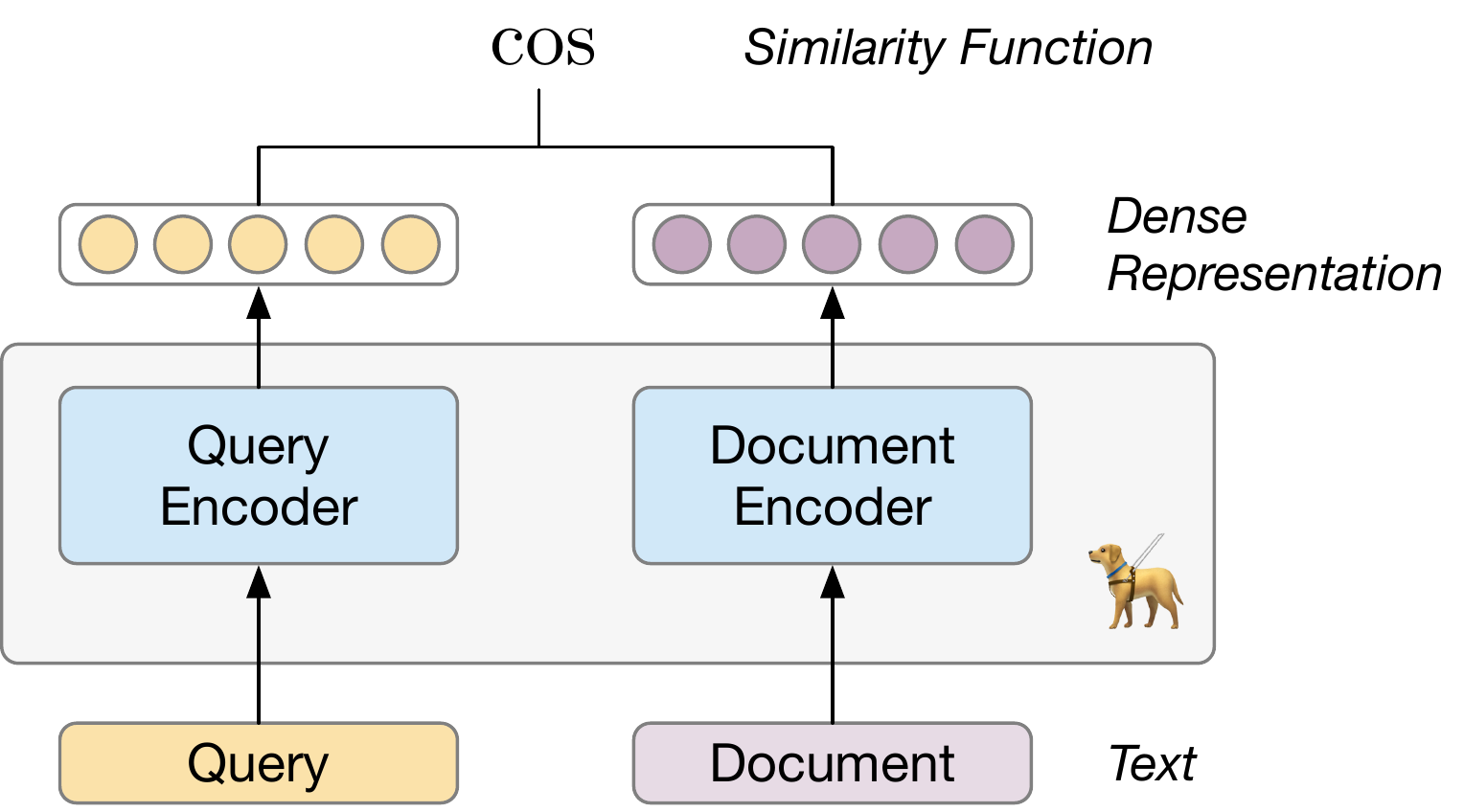}
    \caption{Dual-tower architecture for text retrieval.}
    \label{fig:dualtower}
\end{figure}

\section{Methodology}

\subsection{Dual-Tower Architecture}
\paragraph{Two Encoders} The dual-tower architecture,
as illustrated in Figure~\ref{fig:dualtower}, is widely used in dense retrieval systems~\citep{lee2019latent,dpr,ance}. The dual-tower architecture has a query encoder $E_Q$ and a document encoder $E_D$, which in our work are both BERT-like bidirectional text encoders~\citep{bert}. Compared with cross-attention models~\citep{sbert,gao2020modularized,macavaney2020efficient}, the dual-tower architecture enables pre-indexing and fast approximate nearest neighbor search (to be detailed shortly), thus is popular in production.

\paragraph{Dense Representation} Given an input document (query) $\bm{x}=\{\texttt{[CLS]}, w_1, \ldots, w_l, \texttt{[SEP]}\}$, we use a document (query) encoder $E_D$ ($E_Q$) to encode the input sequence into hidden states $\bm{h}=\{v_\texttt{[CLS]}, v_1, \ldots, v_l, v_\texttt{[SEP]}\}$, where $w_i$ is the $i$-th token; \texttt{[CLS]} and \texttt{[SEP]} are special tokens that mark the start and end of a sentence, respectively. To obtain a dense representation, we use 
mean pooling over hidden states $\bm{h}$ as the representation $\bm{h_x}$ of the input $\bm{x}$. 
Some prior works~\citep{lee2019latent,ict,dpr} use $v_\texttt{[CLS]}$ as the representation for the input $\bm{x}$, but \citet{huang2021whiteningbert} empirically find that applying mean pooling to hidden states $\bm{h}$
outperforms taking $v_\texttt{[CLS]}$ as the representation. 

\paragraph{Similarity Function} After obtaining the representation for both the query $q$ and the document $d$, we use the cosine function as a similarity function to measure the similarity between them:
\begin{equation}
    \mathrm{sim}(q, d) = \frac{E_Q(q) \cdot E_D(d)}{\|E_Q(q)\| \| E_D(d) \|}
    \label{eq:sim}
\end{equation}

\paragraph{Approximate Nearest Neighbor} In practice, for the dual-tower architecture, the documents are encoded offline and their dense representations can be pre-indexed by a fast vector similarity search library (\eg FAISS, \citealp{faiss}). The library can utilize GPU acceleration to perform approximate nearest neighbor (ANN) search in sub-linear time with almost no 
loss in
recall. Thus, compared to a cross-encoder (\ie an encoder that accepts the concatenation of the query and every candidate document), a pre-indexed ANN-based retrieval system is at least 10 times faster (to be detailed in Section \ref{sec:res}).

\subsection{Constructing Positive Instances}
In this section, we first introduce how we build the positive instances with two self-supervised tasks, namely Inverse Cloze Task (ICT) and Dropout as Positive Instance (DaPI).%

\paragraph{Inverse Cloze Task (ICT)} First introduced in \citet{lee2019latent}, ICT is an effective way to pretrain a text retrieval model~\citep{ict}. Given a passage $p$ consisting of sentences $p=\{s_1, \ldots, s_n\}$, we randomly select a sentence $s_k$ as query $q$ and treat its context as document $d=\{s_1, \ldots, s_{k-1}, s_{k+1}, \ldots, s_n\}$. ICT is designed to mimic a text retrieval task where a short query is used to retrieve a longer document which is semantically relevant. Also, unlike some pretraining tasks, \eg Wiki Link Prediction or Body First Selection~\citep{ict}, ICT is fast and does not rely on a specific corpus format (\eg Wikipedia) thus can be scaled to a large multi-source corpus (\eg C4, \citealp{t5}).

\paragraph{Dropout as Positive Instance (DaPI)} DaPI is originally proposed in SimCSE~\citep{gao2021simcse} as a simple strategy for perturbing intermediate representations and thus can serve as data augmentation.\footnote{To avoid confusion with the SimCSE model, we address the dropout strategy as DaPI here.} A similar idea is also presented in \citet{liu2021fast}. We apply a dropout rate of 0.1 to the fully-connected layers and attention probabilities in the Transformer encoders, as in BERT~\citep{bert}. The same input is fed to the encoder twice to obtain two representations, of which one is used as the positive instance of the other.
\citet{gao2021simcse} conduct experiments and conclude that the dropout strategy outperforms all commonly-used discrete perturbation techniques including cropping, word deletion, masked language modeling and synonym replacement. Note that different from SimCSE, we only calculate gradients for one of the two passes. In our experiments, we find that the addition of DaPI only increases the memory use by 2\%, since it mostly reuses the computational graph for the ICT objective.

\subsection{Iterative Contrastive Learning}
\begin{figure*}
     \centering
     \begin{subfigure}[b]{0.45\textwidth}
         \centering
         \includegraphics[width=\textwidth]{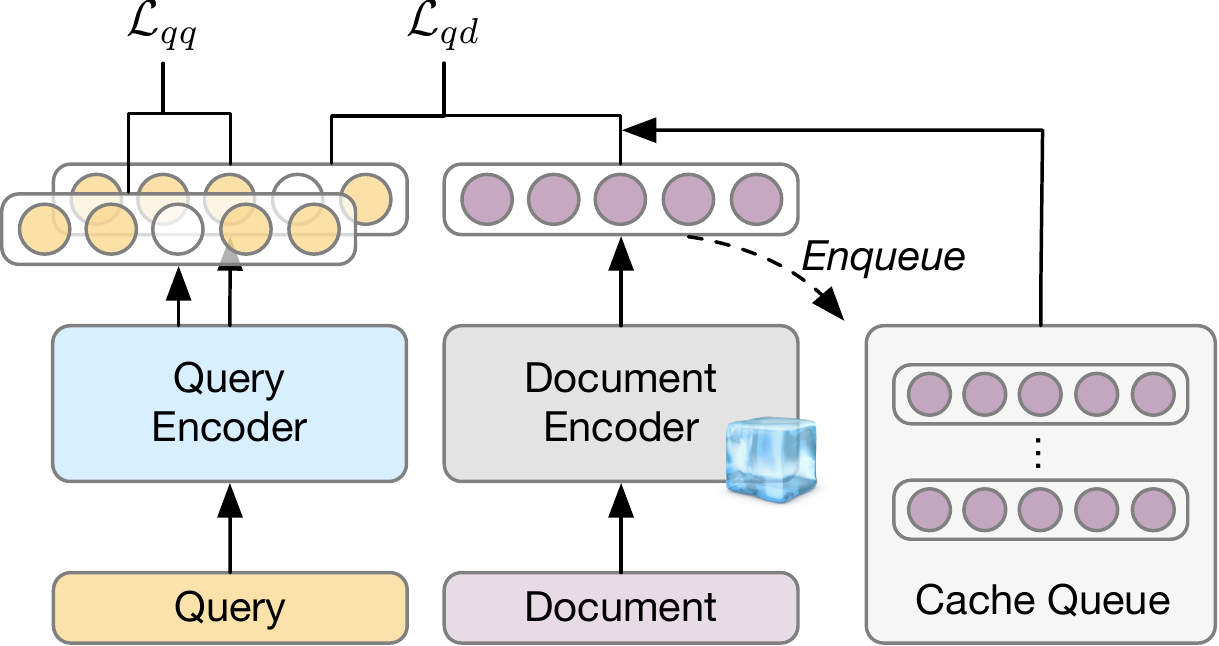}
         \caption{Query encoder training.}
         \label{fig:icolq}
     \end{subfigure}
     \hfill
     \begin{subfigure}[b]{0.45\textwidth}
         \centering
         \includegraphics[width=\textwidth]{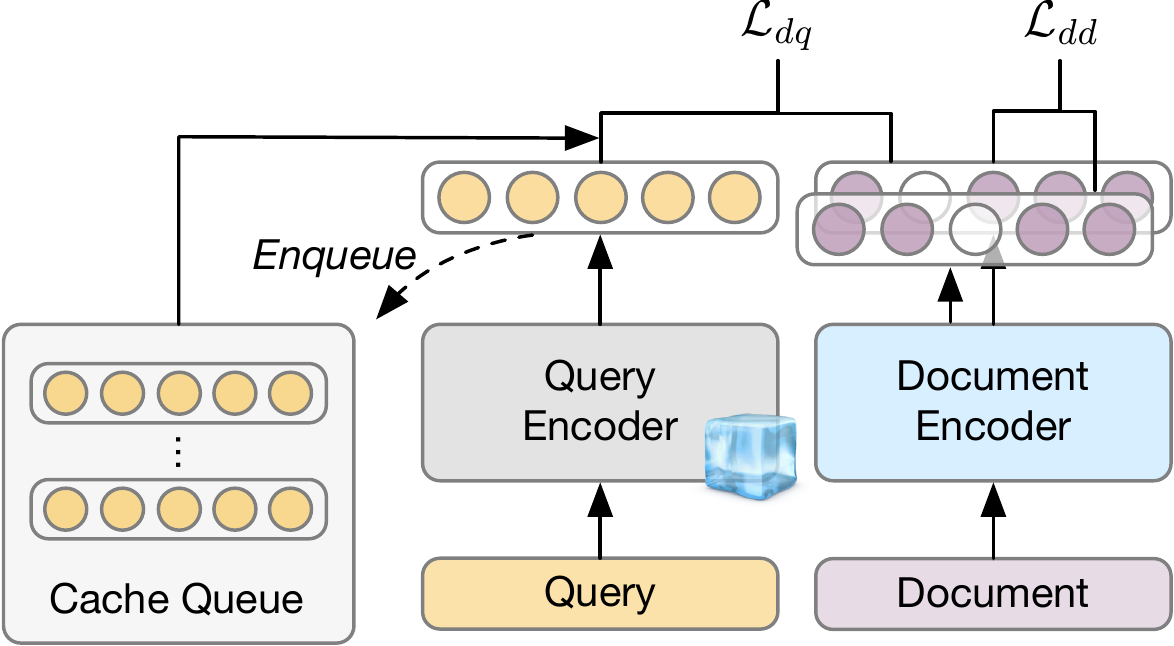}
         \caption{Document encoder training.}
         \label{fig:icold}
     \end{subfigure}
     \caption{Training of \baby with Iterative Contrastive Learning (ICoL). We iteratively train the query encoder and document encoder while freezing the other (marked with an ice cube icon \includegraphics[height=1em]{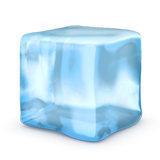}). For $\mathcal{L}_{qd}$ and $\mathcal{L}_{dq}$, we obtain additional negative instances from the cache queue. For each batch of data, we enqueue the representation encoded by the frozen encoder into the cache queue as future negative instances. The cache queue is cleared when switching the encoder to train from one to the other. \label{fig:icol}}
\end{figure*}

Previous studies~\citep{declutr,wu2020clear,gao2021scaling} show that the %
number
of negative instances is critical to the performance of the model. Since the batch size on a single GPU is limited, we propose Iterative Contrastive Learning (ICoL) to mitigate the insufficient memory on a single GPU and allow more negative instances for better performance. We illustrate \baby training in Figure~\ref{fig:icol}.

\paragraph{Iterative Training} We iteratively train the query encoder and document encoder. To be specific, we first arbitrarily select an encoder to start training. Here we assume to start with the query encoder $E_Q$. The training loss consists of two terms. First, we calculate the loss for query-query retrieval with DaPI to optimize the negative log likelihood of the positive instance:
\begin{equation}
\begin{aligned}
     & \mathcal{L}_{qq}(q_i, \{q^+_i, q^-_{i,1}, \ldots, q^-_{i,n}\}) 
    \\ = & -\log \frac{e^{\operatorname{sim}\left(q_{i}, q_{i}^{+}\right)}}{e^{\operatorname{sim}\left(q_{i}, q_{i}^{+}\right)}+\sum_{j=1}^{n} e^{\operatorname{sim}\left(q_{i}, q_{i, j}^{-}\right)}}
\end{aligned}
\end{equation}
where $q_i$ and $q_i^+$ are the same query that are encoded by $E_Q$ with different dropout masks; $\{q^-_{i,1}, ..., q^-_{i,n}\}$ is a set of randomly sampled negative instances; $\operatorname{sim}(\cdot,\cdot)$ is the cosine similarity function defined in Equation~\ref{eq:sim}.

The second term is to retrieve the corresponding document $d^+_i$ with the query $q_i$, where $q_i$ and $d^+_i$ are a pair constructed with ICT. Similarly, we optimize the negative log likelihood of the positive instance by:
\begin{equation}
\begin{aligned}
     & \mathcal{L}_{qd}(q_i, \{d^+_i, d^-_{i,1}, \ldots, d^-_{i,n}, d^-_{\mathcal{Q}, 1}, \ldots, d^-_{\mathcal{Q}, |\mathcal{Q}|}\}) 
    \\ = & -\log \frac{e^{\operatorname{sim}\left(q_{i}, d_{i}^{+}\right)}}{\splitfrac{e^{\operatorname{sim}\left(q_{i}, d_{i}^{+}\right)}+\sum_{j=1}^{n} e^{\operatorname{sim}\left(q_{i}, d_{i, j}^{-}\right)}}{+ \sum_{k=1}^{|\mathcal{Q}|} e^{\operatorname{sim}\left(q_{i}, d_{\mathcal{Q}, k}^{-}\right)}}}
\end{aligned}
\end{equation}
where $\{d^-_{i,1}, ..., d^-_{i,n}\}$ is a set of freshly sampled documents that are encoded at the current step $i$; $\{d^-_{\mathcal{Q},1}, ..., d^-_{\mathcal{Q}, |\mathcal{Q}|}\}$ is a set of representations that are currently stored in the cache queue $\mathcal{Q}$. Then, we optimize the sum of the two losses with a weight coefficient $\lambda$:
\begin{equation}
    \mathcal{L}_q = \mathcal{L}_{qd} + \lambda \mathcal{L}_ {qq}
\end{equation}
Note that the query $q_i$ only needs to be encoded once and can be used for calculation of both $\mathcal{L}_{qd}$ and $\mathcal{L}_{qq}$. 

After a predefined number %
of steps, the $E_Q$ becomes frozen as the training for $E_D$ starts. Similarly, for $d_i$, a document encoded by $E_D$, we have the training objective:
\begin{equation}
\begin{aligned}
     & \mathcal{L}_{dd}(d_i, \{d^+_i, d^-_{i,1}, \ldots, d^-_{i,n}\}) 
    \\ = & -\log \frac{e^{\operatorname{sim}\left(d_{i}, d_{i}^{+}\right)}}{e^{\operatorname{sim}\left(d_{i}, d_{i}^{+}\right)}+\sum_{j=1}^{n} e^{\operatorname{sim}\left(d_{i}, d_{i, j}^{-}\right)}}
\end{aligned}
\end{equation}
\begin{equation}
\begin{aligned}
     & \mathcal{L}_{dq}(d_i, \{q^+_i, q^-_{i,1}, \ldots, q^-_{i,n}, q^-_{\mathcal{Q}, 1}, \ldots, q^-_{\mathcal{Q}, |\mathcal{Q}|}\}) 
    \\ = & -\log \frac{e^{\operatorname{sim}\left(d_{i}, q_{i}^{+}\right)}}{\splitfrac{e^{\operatorname{sim}\left(d_{i}, q_{i}^{+}\right)}+\sum_{j=1}^{n} e^{\operatorname{sim}\left(d_{i}, q_{i, j}^{-}\right)}}{+ \sum_{k=1}^{|\mathcal{Q}|} e^{\operatorname{sim}\left(d_{i}, q_{\mathcal{Q}, k}^{-}\right)}}}
    \end{aligned}
\end{equation}
\begin{equation}
    \mathcal{L}_d = \mathcal{L}_{dq} + \lambda \mathcal{L}_ {dd}
\end{equation}
where $d^+_i$ and $q^+_i$ are positive instances constructed by DaPI and ICT, respectively; $\{d^-_{i,1}, \ldots, d^-_{i,n}\}$ is a set of randomly sampled document negatives; $\{q^-_{i,1}, \ldots, q^-_{i,n}\}$ is a set of freshly sampled queries encoded at step $i$; $\{q^-_{\mathcal{Q}, 1}, \ldots, q^-_{\mathcal{Q}, |\mathcal{Q}|}\}$ are the cached query representations. To speed up training, we apply the in-batch negatives technique~\citep{yih2011learning,henderson2017efficient,gillick2019learning} that can reuse computation and train $b$ queries/documents in a mini-batch simultaneously. 

\paragraph{Cache Mechanism} To enlarge the size of negative instances, we maintain a cache queue $\mathcal{Q}$ that stores previously encoded representations that can serve as negative instances for the current step, extending an earlier study~ \citep{wu2018unsupervised}.
Our cache queue is implemented as first-in-first-out (FIFO) with a maximum capacity $m$, which is a hyperparameter set based on the GPU memory size. When training with multiple GPUs, $\mathcal{Q}$ can be shared across GPUs. Since the representations in the queue are encoded with a frozen encoder and thus do not require gradients, $m$ can be set large to
supplement the numbers of negative instances. When $\mathcal{Q}$ is full, the earliest cached representations will be dequeued. When we switch the training from one encoder to the other, the queue will be cleared to ensure that all representations in $\mathcal{Q}$ lie in the same hidden space and are encoded with the currently frozen encoder. 

\paragraph{ICoL vs.~MoCo}
Previously, similar to our method, MoCo~\citep{moco} exploits a queue for storing encoded representations.
Specifically, MoCo consists of a slow encoder and a fast encoder to encode queries and documents, respectively. The slow encoder is updated as a slow moving average of the fast encoder to reduce inconsistency of encoded document representations between training steps. A queue is maintained to allow the encoded document representations to be reused in later steps as negative instances.

However, we argue there are a two 
limitations that make
MoCo not ideal for training a text retrieval model: (1) As pointed out by \citet{xmoco}, unlike the image matching task in the original paper of MoCo, in text retrieval, the queries and documents are distinct from each other thus not interchangeable. \citet{xmoco} propose xMoCo, which incorporates two sets of slow and fast encoders, as a simple fix for this flaw. (2) The cached representations are in different hidden spaces. Although the fast encoders in both MoCo and xMoCo are updated with momentum, the already-encoded representations in the queue will never be updated. This creates a semantic mismatch between newly encoded and cached old representations and creates noise during training. In ICoL, all representations used for contrastive learning are aligned in the same hidden space.
Besides, ICoL is more flexible than xMoCo since it does not introduce additional fast encoders and even the weights of its query encoder and document encoder can be shared. We conduct experiments to compare ICoL with MoCo and xMoCo in Section \ref{sec:icol}.

\subsection{Lexicon-Enhanced Dense Retrieval}
Although dense retrieval achieves state-of-the-art performance, its performance significantly degenerates on out-of-domain data~\citep{thakur2021beir}. On the other hand, BM25~\citep{bm25} demonstrates %
good performance without training.
Early attempts at combining lexical match with dense retrieval often formulate it to a re-ranking task~\citep{msmarco}. First, BM25 is used to recall the top-$k$ documents from the corpus. Then, a cross-encoder is applied to further re-rank candidate documents.
Recently, COIL~\citep{coil} highlights the importance of lexical match and %
incorporates
exact lexical matching into dense retrieval. Different from these works, we propose a fast and effective way, namely Lexicon-Enhanced Dense Retrieval (LEDR) to enhance dense retrieval with BM25. The similarity score of BM25 is defined as:
\begin{equation}
\begin{aligned}
\operatorname{BM25}(q, d)&=\sum_{t \in q \cap d} \operatorname{IDF}(t) h_{q}(q, t) h_{d}(d, t) \\
h_{q}(q, t)&=\frac{\operatorname{TF}_{t, q}\left(1+k_{2}\right)}{\operatorname{TF}_{t, q}+k_{2}} \\
h_{d}(d, t)&=\frac{\operatorname{TF}_{t, d}\left(1+k_{1}\right)}{\operatorname{TF}_{t, d}+k_{1}\left(1-b+b \frac{|d|}{\text { avgdl }}\right)}
\end{aligned}
\end{equation}
where $\operatorname{TF}_{t,d}$ and $\operatorname{TF}_{t,q}$ refer to term frequency of term $t$ in document $d$ and query $q$, respectively; $\operatorname{IDF}(t)$ is the inverse document frequency; $b$, $k_1$ and $k_2$ are hyperparameters.
For inference, we simply multiply the BM25 score with the similarity score for dense retrieval:%
\begin{equation}
    \mathit{score}(q,d) = \operatorname{sim}(q,d) \times \operatorname{BM25}(q,d)
\end{equation}
In this way, we consider both lexical and semantic matching. This combination makes \baby more robust on unseen data in zero-shot learning. %

 \begin{table*}[htbp]
  \centering
  \resizebox{1\linewidth}{!}{
    \begin{tabular}{clccccccccc}
    \toprule
    \multicolumn{2}{c}{\multirow{2}{*}{\textbf{Model}}} & \multicolumn{4}{c}{\textbf{Dense Retrieval}} & \textbf{Lexical} & \textbf{Late Interaction} & \textbf{Re-ranking} & \multicolumn{2}{c}{\textbf{Lexicon-Enhanced Dense}} \\
    \cmidrule{3-6} \cmidrule{10-11}
    & & DPR   & ANCE  & GenQ  & TAS-B & BM25$^\dagger$ & ColBERT & BM25 + CE & LaPraDoR$^\dagger$ & \baby FT \\
    \midrule
    \textbf{Encoding} & Qry/s (GPU/CPU) & 4000/170 & 4000/170 & 4000/170 & 7000/350 & -     & 4000/170 & 7000/350 & 7000/350 & 7000/350 \\
    \textbf{Speed} & Doc/s (GPU/CPU) & 540/30 & 540/30 & 540/30 & 1100/70 & -     & 540/30 & 1100/70 & 1100/70 & 1100/70 \\
    \midrule
    \textbf{Index size} & & 3 GB  & 3 GB  & 3 GB  & 3 GB  & 0.4 GB & 20 GB & 0.4 GB & 3.4 GB & 3.4 GB \\
    \midrule
    \textbf{Retrieval} & GPU & 19 ms & 20 ms & 14 ms & 14 ms & -     & 350 ms & 450 ms & 20 ms & 20 ms \\
   \textbf{Latency} & CPU & 230 ms & 275 ms & 125 ms & 125 ms & 20 ms & -     & 6100 ms & 145 ms & 145 ms \\
    \midrule
    \textbf{MS-MARCO} & nDCG@10 & 0.177 & 0.388 & 0.408 & 0.408 & 0.228 & 0.401 & 0.413 & 0.262 & 0.366 \\
    \midrule
    \multirow{19}[2]{*}{\textbf{Zero-shot}} & TREC-COVID & 0.332 & 0.654 & 0.619 & 0.481 & 0.656 & 0.677 & 0.757 & 0.728 & \textbf{0.779} \\
          \multirow{19}[2]{*}{(nDCG@10)} & BIOASQ & 0.127 & 0.306 & 0.398 & 0.383 & 0.465 & 0.474 & \textbf{0.523} & 0.500 & 0.511 \\
          & NFCorpus & 0.189 & 0.237 & 0.319 & 0.319 & 0.325 & 0.305 & \textbf{0.350} & 0.346 & 0.347 \\
          & NQ    & 0.474 & 0.446 & 0.358 & 0.463 & 0.329 & 0.524 & \textbf{0.533} & 0.359 & 0.479 \\
          & HotpotQA & 0.391 & 0.456 & 0.534 & 0.584 & 0.603 & 0.593 & \textbf{0.707} & 0.625 & 0.666 \\
          & FiQA  & 0.112 & 0.295 & 0.308 & 0.300 & 0.236 & 0.317 & \textbf{0.347} & 0.317 & 0.343 \\
          & Signal-1M & 0.155 & 0.249 & 0.281 & 0.289 & 0.330 & 0.274 & 0.338 & 0.343 & \textbf{0.344} \\
          & TREC-NEWS & 0.161 & 0.382 & 0.396 & 0.377 & 0.398 & 0.393 & 0.431 & 0.470 & \textbf{0.480} \\
          & Robust04 & 0.252 & 0.392 & 0.362 & 0.427 & 0.408 & 0.391 & 0.475 & \textbf{0.490} & 0.484 \\
          & ArguAna & 0.175 & 0.415 & 0.493 & 0.429 & 0.315 & 0.232 & 0.311 & 0.507 & \textbf{0.508} \\
          & Touche-2020 & 0.131 & 0.240 & 0.182 & 0.162 & 0.367 & 0.202 & 0.271 & 0.322 & \textbf{0.333} \\
          & CQADupStack & 0.153 & 0.296 & 0.347 & 0.314 & 0.299 & 0.350 & \textbf{0.370} & 0.222 & 0.290 \\
          & Quora & 0.248 & 0.852 & 0.830 & 0.835 & 0.789 & 0.854 & 0.825 & 0.863 & \textbf{0.875} \\
          & DBPedia & 0.263 & 0.281 & 0.328 & 0.384 & 0.313 & 0.392 & \textbf{0.409} & 0.361 & 0.391 \\
          & SCIDOCS & 0.077 & 0.122 & 0.143 & 0.149 & 0.158 & 0.145 & 0.166 & \textbf{0.185} & 0.184 \\
          & FEVER & 0.562 & 0.669 & 0.669 & 0.700 & 0.753 & 0.771 & \textbf{0.819} & 0.671 & 0.763 \\
          & Climate-FEVER & 0.148 & 0.198 & 0.175 & 0.228 & 0.213 & 0.184 & 0.253 & 0.228 & \textbf{0.261} \\
          & SciFact & 0.318 & 0.507 & 0.644 & 0.643 & 0.665 & 0.671 & 0.688 & \textbf{0.697} & 0.687 \\
          \cmidrule{2-11}
          & Avg.  & 0.237 & 0.389 & 0.410 & 0.415 & 0.423 & 0.431 & 0.476 & 0.457 & \textbf{0.485} \\
    \bottomrule
    \end{tabular}%
    }
  \caption{Experimental results on the BEIR benchmark~\citep{thakur2021beir}. The estimated average retrieval latency and index sizes are for a single query in DBPedia. The encoding speed is reported on a 8-core Intel Xeon Platinum 8168 CPU @ 2.70GHz and a single Nvidia V100 GPU, respectively. ``\baby FT'' is a \baby model fine-tuned on MS-MARCO with the official BEIR training script. $^\dagger$Unsupervised method. }
  \label{tab:main}%
\end{table*}%

\section{Experiments}
\subsection{Experimental Setting}
\paragraph{Benchmark} We use BEIR~\citep{thakur2021beir}, a recently released benchmark for zero-shot evaluation of information retrieval models. BEIR includes 18 heterogeneous datasets, focusing on evaluating a retrieval system that works across different domains (bio-medical, scientific, news, social media, etc.). The benchmark uses Normalized Discounted Cumulative Gain (nDCG) \cite{jarvelin2002cumulated} as the evaluation metric, which is a measure of ranking quality and often used to measure effectiveness of search algorithms or retrieval models.
Details of the BEIR benchmark and the evaluation metric are included in Appendix~\ref{sec:beir}. 

\paragraph{Model Settings} In our preliminary experiments on Wikipedia (see Table~\ref{tab:icol}), we find that sharing weights between the query encoder $E_Q$ and document encoder $E_D$ has no
negative effect on downstream performance.
For weight sharing between $E_Q$ and $E_D$, we simply copy the weights of $E_Q$ to $E_D$ when switching to training of $E_D$, vice versa. This design eliminates nearly half of the parameters. An additional benefit is that weight sharing makes the encoder versatile to handle not only query-document retrieval, but also query-query and document-document retrieval.

 In our preliminary experiments on Wikipedia, we observed a diminishing return when increasing the model size from 6 layers to 12 layers, or 24 layers.
 Thus, we initialize our encoder with the 6-layer DistilBERT~\citep{sanh2019distilbert}, which has $\sim$67M parameters. %
 For BM25, we use the implementation and default settings of Elastic Search\footnote{\url{https://github.com/elastic/elasticsearch}}. BM25 scores after the top 1,000 retrieved text are set to 0 to save computation.

\paragraph{Training Details} For pretraining, we optimize the model with the AdamW optimizer with a learning rate of 2e-4. The model is trained with 16 Nvidia V100 32GB GPUs with FP16 mixed precision training. The batch size for each GPU is set to 256. The maximum lengths set for queries and documents are 64 and 350, respectively. Training 
switches
between $E_Q$ and $E_D$ every 100 steps. The cache queue has a maximum capacity $m$ of 100k. The loss weight hyperparameter $\lambda$ is fixed to 1. For our main results, we train \baby on C4~\citep{t5} for 1M steps, which takes about 400 hours. For
the ablation study, since training on C4 is very costly, we train \baby on Wikipedia\footnote{\url{https://huggingface.co/datasets/wikipedia}} for 100k steps. When calculating the loss, we apply a re-scaling trick of multiplying the cosine similarity score by 20 for better optimization~\citep{thakur2021beir}. Our implementation of \baby is based on Hugging Face Transformers~\citep{hf} and Datasets~\citep{hfdatasets}.

We test \baby under two settings: \textbf{(1) No supervised data at all.} We directly use the pretrained model for zero-shot retrieval on BEIR. \textbf{(2) Fine-tuning on MS-MARCO}~\citep{msmarco} and zero-shot transfer to the other datasets. This is the original setting for BEIR. We use BEIR's official script\footnote{\url{https://github.com/UKPLab/beir/blob/main/examples/retrieval/training/train_msmarco_v3.py}} to fine-tune \baby. The batch size is set to 75 per GPU and the learning rate is 2e-5. 

\paragraph{Baselines} For dense retrieval, we compare our model to the dual-tower models: DPR~\cite{dpr}, ANCE~\cite{ance}, TAS-B~\cite{tasb} and GenQ~\cite{thakur2021beir}. For lexical matching, we use the BM25 results reported in \citet{thakur2021beir}. We also consider a late interaction baseline \textbf{ColBERT}~\cite{colbert}. The model computes multiple contextualized embeddings for each token of queries and documents, and then 
maximizes a
similarity function to retrieve relevant documents. For re-ranking, we use the \textbf{BM25+CE} baseline implemented in \citet{thakur2021beir} that uses BM25 to retrieve top-100 documents and a cross-encoder model to further re-rank.
As shown in Table~\ref{tab:main}, the latency for both lexical and dense retrieval is low whereas re-ranking introduces significantly higher latency, with late-interaction in-between. Details of the baselines can be found in Appendix~\ref{sec:baseline}. 

\begin{table}[t]
  \centering
  \resizebox{1\columnwidth}{!}{
    \begin{tabular}{clccccc}
    \toprule
    \multicolumn{2}{c}{\multirow{2}{*}{\textbf{Model}}} & In-Batch & \multirow{2}{*}{MoCo} & \multirow{2}{*}{xMoCo} & \multirow{2}{*}{ICoL} & ICoL\\
    & & (shared) & & & & (shared) \\
    \midrule
    \textbf{\#Encoder} & & 1 & 2 & 4 & 2 & 1 \\
    \midrule
    \multicolumn{1}{l}{\textbf{MS-MARCO}} & nDCG@10 & 0.255 & 0.222 & 0.255 & 0.255 & \textbf{0.262} \\
    \midrule
    \multirow{19}[2]{*}{\textbf{Zero-shot}} & TREC-COVID & 0.705 & 0.537 & \textbf{0.724} & 0.706 & 0.710 \\
        \multirow{19}[2]{*}{(nDCG@10)} & BIOASQ & 0.451 & 0.260  & 0.423 & \textbf{0.468} & 0.459 \\
          & NFCorpus & 0.315 & 0.271 & 0.312 & \textbf{0.317} & 0.314 \\
          & NQ & 0.332 & 0.279 & \textbf{0.355} & \textbf{0.355} & 0.351  \\
          & HotpotQA & 0.599 & 0.552 & 0.584 & 0.598  & \textbf{0.610}\\
          & FiQA  & 0.213 & 0.156 & 0.242 & \textbf{0.256} & 0.251 \\
          & Signal-1M & 0.329 & 0.307 & 0.323 & 0.327 & \textbf{0.335} \\
          & TREC-NEWS & 0.441 & 0.405 & 0.441 & 0.444 & \textbf{0.445}\\
          & Robust04 & 0.419 & 0.439 & 0.439 & 0.465 & \textbf{0.470}\\
          & ArguAna & 0.477 & 0.465 & 0.491 & 0.496 & \textbf{0.503} \\
          & Touche-2020 & 0.302 & 0.261 & 0.330 & \textbf{0.331} & 0.328 \\
          & CQADupStack & 0.109 & 0.052 & 0.118 & 0.132 & \textbf{0.140} \\
          & Quora & 0.832 & 0.834 & 0.822 & 0.828 & \textbf{0.839} \\
          & DBPedia & 0.349 & 0.318 & 0.359 & \textbf{0.374} & 0.364 \\
          & SCIDOCS & 0.173 & 0.154 & 0.170  & 0.173 & \textbf{0.178}\\
          & FEVER & 0.537 & 0.540  & 0.651 & \textbf{0.686} & 0.653 \\
          & Climate-FEVER & 0.206 & 0.183 & \textbf{0.244} & 0.242 & 0.242 \\
          & SciFact & 0.660 & 0.659 & 0.667 & 0.683 & \textbf{0.689} \\
          \cmidrule{2-7}
          & Avg. & 0.414 & 0.371 & 0.428 & \textbf{0.438} & \textbf{0.438} \\
    \bottomrule
    \end{tabular}%
    }
    \caption{Comparison of different methods for contrastive learning. The models are trained on Wikipedia.}
  \label{tab:icol}%
\end{table}%

\begin{table}[t]
  \centering
  \resizebox{1\columnwidth}{!}{
    \begin{tabular}{lcc|cccc}
    \toprule
    \multicolumn{1}{c}{\multirow{2}[2]{*}{Model}} & \multicolumn{2}{c}{\baby} & \multicolumn{4}{c}{\baby FT} \\
    \cmidrule(lr){2-3} \cmidrule(lr){4-7}
          & \multicolumn{1}{l}{Full} & w/o LEDR & Full & w/o LEDR & w/o PT & w/o LEDR \& PT \\
    \midrule
    TREC-COVID & \textbf{0.728} & 0.227 & \textbf{0.779} & 0.492 & 0.735 & 0.482 \\
    BIOASQ & \textbf{0.500} & 0.205 & \textbf{0.511} & 0.308 & 0.489 & 0.281 \\
    NFCorpus & \textbf{0.346} & 0.311 & \textbf{0.347} & 0.335 & 0.323 & 0.267 \\
    NQ    & \textbf{0.359} & 0.181 & \textbf{0.479} & 0.473 & 0.454 & 0.443 \\
    HotpotQA & \textbf{0.625} & 0.303 & \textbf{0.666} & 0.495 & 0.642 & 0.484 \\
    FiQA  & \textbf{0.317} & 0.203 & \textbf{0.343} & 0.314 & 0.308 & 0.245 \\
    Signal-1M & \textbf{0.343} & 0.186 & 0.344 & 0.231 & \textbf{0.354} & 0.247 \\
    TREC-NEWS & \textbf{0.470} & 0.345 & \textbf{0.480} & 0.374 & 0.449 & 0.350 \\
    Robust04 & \textbf{0.490} & 0.319 & \textbf{0.484} & 0.368 & 0.459 & 0.332 \\
    ArguAna & \textbf{0.507} & 0.459 & \textbf{0.508} & 0.469 & 0.495 & 0.412 \\
    Touche-2020 & \textbf{0.322} & 0.094 & 0.333 & 0.182 & \textbf{0.346} & 0.156 \\
    CQADupStack & \textbf{0.222} & 0.220 & 0.290 & 0.288 & \textbf{0.306} & 0.250 \\
    Quora & \textbf{0.863} & 0.787 & \textbf{0.875} & 0.847 & 0.867 & 0.840 \\
    DBPedia & \textbf{0.361} & 0.250 & \textbf{0.391} & 0.338 & 0.384 & 0.303 \\
    SCIDOCS & \textbf{0.185} & 0.133 & \textbf{0.184} & 0.155 & 0.173 & 0.127 \\
    FEVER & \textbf{0.671} & 0.368 & \textbf{0.763} & 0.646 & 0.750 & 0.664 \\
    Climate-FEVER & \textbf{0.228} & 0.138 & \textbf{0.261} & 0.209 & 0.247 & 0.206 \\
    SciFact & \textbf{0.697} & 0.555 & \textbf{0.687} & 0.599 & 0.678 & 0.529 \\
    \midrule
    Avg.  & \textbf{0.457} & 0.294 & \textbf{0.485} & 0.396 & 0.470 & 0.368 \\
    \bottomrule
    \end{tabular}%
    }
  \caption{Effect of pretraining (PT) and Lexicon-Enhanced Dense Retrieval (LEDR). Pretraining is on C4. The results of ``w/o PT'' directly use DistilBERT~\citep{sanh2019distilbert} for fine-tuning, which is also used to initialize our model.}
  \label{tab:ptledr}%
\end{table}%

\begin{table}[t]
  \centering
  \resizebox{0.75\columnwidth}{!}{
    \begin{tabular}{lccc}
    \toprule
    \multicolumn{1}{c}{Model} & \baby & w/o DaPI & w/o ICT \\
    \midrule
    TREC-COVID & 0.710 & \textbf{0.714} & 0.612 \\
    BIOASQ & \textbf{0.459} & 0.457 &0.270 \\
    NFCorpus & 0.314 & \textbf{0.316} & 0.257\\
    NQ    & 0.351 & \textbf{0.353} & 0.221 \\
    HotpotQA & \textbf{0.610} & 0.608 & 0.431 \\
    FiQA  & \textbf{0.251} & 0.247 &0.145 \\
    Signal-1M & \textbf{0.335} & 0.330 & 0.306 \\
    TREC-NEWS & 0.445 & \textbf{0.448} & 0.336 \\
    Robust04 & \textbf{0.470} & 0.458 & 0.307 \\
    ArguAna & \textbf{0.503} & 0.497 &0.389 \\
    Touche-2020 & \textbf{0.328} & 0.310 &0.248 \\
    CQADupStack & \textbf{0.140} & 0.137 &0.064 \\
    Quora & \textbf{0.839} & 0.839 &0.774 \\
    DBPedia & \textbf{0.364} & 0.363 & 0.242 \\
    SCIDOCS & \textbf{0.178} & 0.173 & 0.113 \\
    FEVER & \textbf{0.653} & 0.639 & 0.376 \\
    Climate-FEVER & \textbf{0.242} & 0.231 & 0.118 \\
    SciFact & 0.689 & \textbf{0.690} & 0.533 \\
    \midrule
    Avg.  & \textbf{0.438} & 0.434&0.319 \\
    \bottomrule
    \end{tabular}%
    }
  \caption{Effect of ICT and DaPI in the loss function. The ``w/o ICT'' variant is equal to the original SimCSE approach~\citep{gao2021simcse}. The pretraining is on Wikipedia.}
  \label{tab:loss}%
\end{table}%

\subsection{Experimental Results}
\label{sec:res}
We list the results of \baby on the BEIR benchmark in Table~\ref{tab:main}.
Our model achieves state-of-the-art performance on BEIR to date (November 15, 2021).
Without any supervised data, \baby outperforms the previous state-of-the-art for zero-shot dense retrieval, TAS-B~\citep{tasb}, on 13 tasks (out of 18) of BEIR with an average advantage of $0.042$, though TAS-B applies additional query clustering and knowledge distillation.
When further fine-tuned on MS-MARCO, \baby can outperform all baselines, including late interaction and re-ranking, whose latency on GPU is 17.5$\times$ and 22.5$\times$ higher than our method. Compared to dense retrieval, we only add 0.4 GB of BM25 indices and almost no additional latency. %

\subsubsection{Effect of Iterative Contrastive Learning}
\label{sec:icol}
We set a baseline that only uses in-batch negatives and compare our proposed Iterative Contrastive Learning (ICoL) to MoCo~\citep{moco} and xMoCo~\citep{xmoco} for training \baby on Wikipedia in Table \ref{tab:icol}. The aforementioned two flaws of MoCo hinder its performance and lead to a performance drop instead of an improvement. 
In contrast,
our ICoL approach outperforms the in-batch baseline on all datasets. It also beats the competitive MoCo variant for text retrieval, xMoCo, on 15 out of 18 tasks. ICoL only uses two encoders (which can be further shared) which can alleviate the GPU memory problem and thus can fit 
more in-batch negatives. Meanwhile, MoCo uses two encoders and xMoCo uses four (two sets of MoCo's encoders). Moreover, we observe no performance drop on average if we share the encoder between query and document (as we do when training \baby on C4). Thus, we can eliminate half of the parameters by simply sharing the encoder.

\subsubsection{Effect of Pretraining and Lexicon-Enhanced Dense Retrieval}
We conduct an ablation study for both pretraining and Lexicon-Enhanced Dense Retrieval to verify the effectiveness of these designs. As shown in Table \ref{tab:ptledr}, Lexicon-Enhanced Dense Retrieval (LEDR) improves performance of dense retrieval on most tasks for both fully unsupervised and fine-tuned \baby. 
Furthermore, as illustrated in Table~\ref{tab:loss}, we test the effectiveness of the two components in our loss function. We can see that both ICT and DaPI significantly contribute to the performance of our model ($p<0.01$) while ICT has a large impact on the final performance.

\begin{figure}[t]
    \centering
    \includegraphics[width=0.97\linewidth]{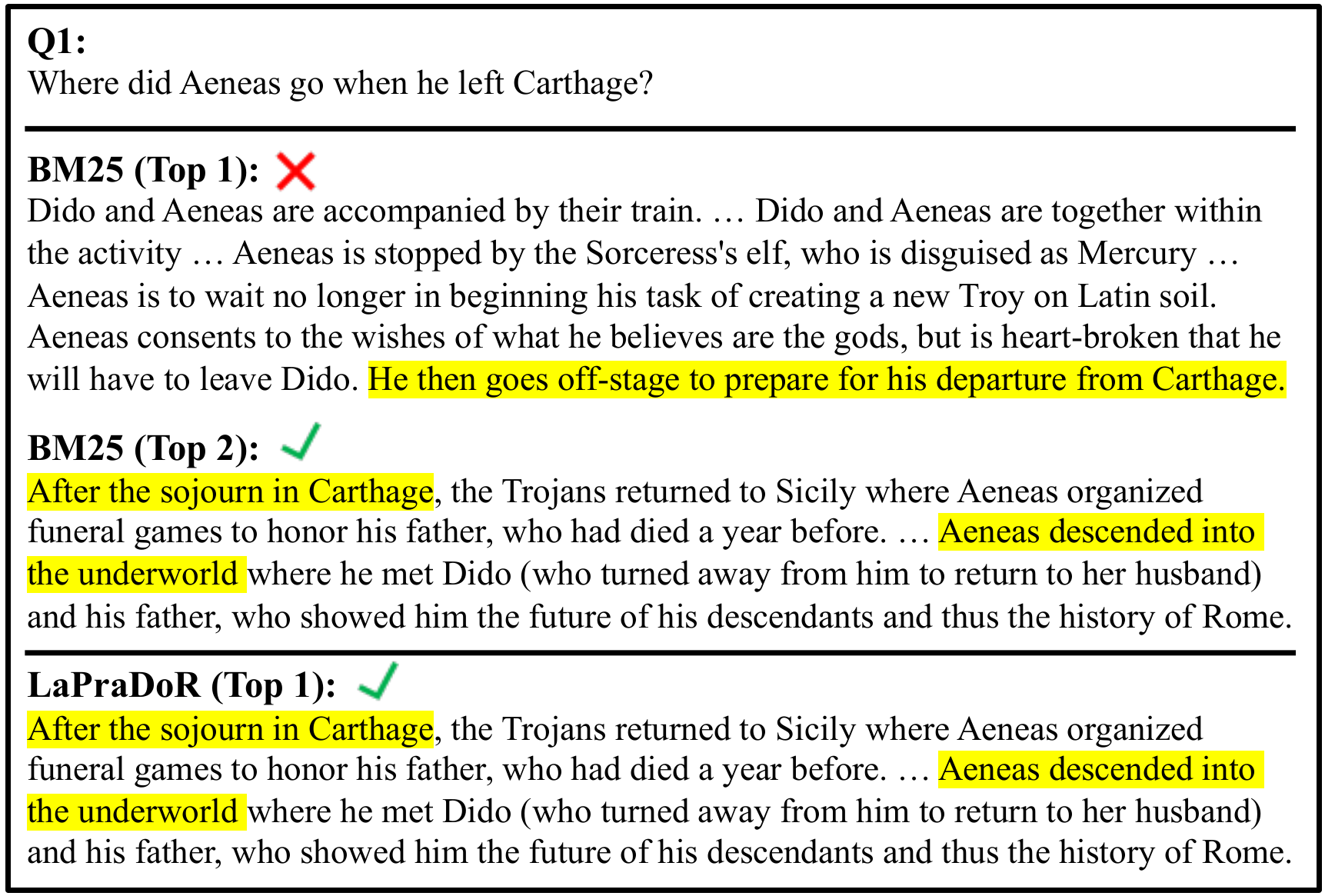}
    \includegraphics[width=0.97\linewidth]{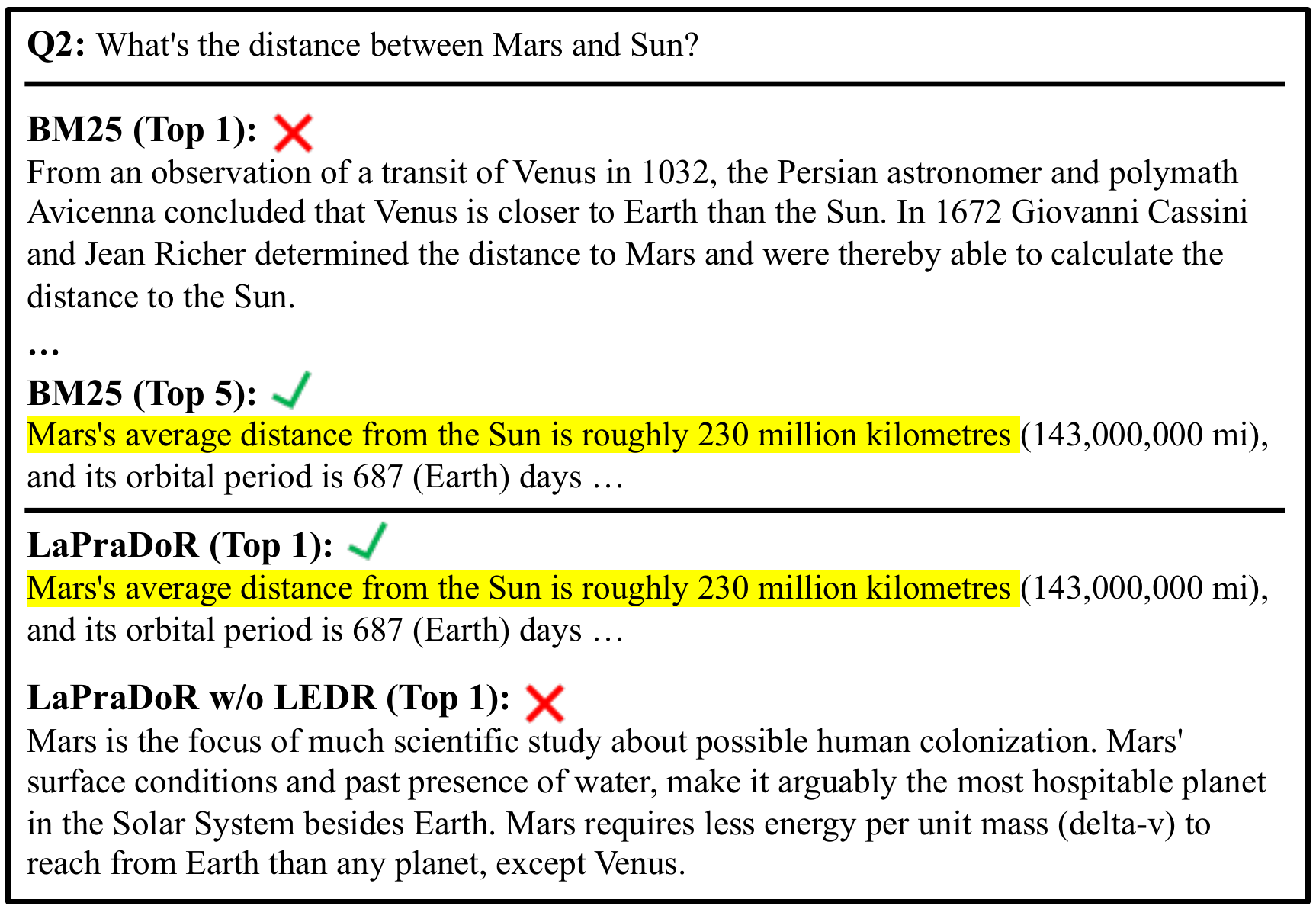}
    \caption{Examples from the NQ dataset~\citep{47761}. The key clues are highlighted.}
    \label{fig:case}
\end{figure}

\subsection{Case Study}
\label{sec:case}
We conduct a case study to intuitively demonstrate the effectiveness of \baby. As shown in Figure~\ref{fig:case}, for Q1, the lexical method (\ie BM25) can successfully find the corresponding document in its top-2 retrieved results. However, due to lower lexical overlap, the score of the ground truth is lower than that of the first document. Although the phrase \textit{``prepare for his departure''} in the first document indicates that \textit{Aeneas has not left Carchage yet} and provides strong evidence that this document is incorrect, BM25 fails to correctly rank the ground truth due to its lack of ability in semantic matching. By incorporating both lexical and semantic matching, \baby can successfully retrieve the ground truth. 

For Q2, with the powerful semantic matching, \baby successfully retrieves the ground truth whereas BM25 fails to distinguish among the documents that contain both the keywords \textit{Mars} and \textit{Sun}. On the other hand, after removing lexical matching, \baby without LEDR suffers from noise: the key entity \textit{Sun} does not appear in its top-1 retrieved document. LEDR helps filter out such noise and allows the dense retriever to focus on fine-grained semantic matching. Please find more cases from other datasets on Appendix \ref{appendix_case}.

\section{Conclusion and Future Work}
In this paper, we introduce \baby, an unsupervised pretrained dense retriever that achieves state-of-the-art performance on the zero-shot text retrieval benchmark BEIR. We propose Iterative Contrastive Learning (ICoL) for efficiently training \baby and Lexicon-Enhanced Dense Retrieval (LEDR) to combine lexical matching with \baby. Our experiments verify the effectiveness of both ICoL and LEDR, shedding light on a new paradigm for unsupervised text retrieval.
For future work, we plan to extend unsupervised \baby to multilingual and multi-modal retrieval.

\section*{Broader Impact}
\paragraph{Ethical Concerns} \baby is trained with web-crawled data, which may contain inappropriate content. However, due to the nature of text retrieval, our retriever has lower ethical risk compared to a %
generative auto-regressive language model~\citep{parrot}. Meanwhile, our unsupervised retrieval model enables high-performance text retrieval for low-resource languages where there is no supervised query-document dataset. This contributes to equality and diversity of language technology.

\paragraph{Carbon Footprint} To conduct all experiments in this paper, we estimate to have consumed 3,840 kWh of electricity and emitted 1,420.8 kg (3,132.3 lbs) of CO$_2$. All emitted carbon dioxide has already been offset by the cloud service provider.
\section*{Acknowledgments}
We would like to thank the anonymous reviewers for their insightful comments. We would like to thank the authors of BEIR~\cite{thakur2021beir}, Nandan Thakur and Nils Reimers, for their support. Canwen wants to thank Minghua Liu's Labrador, \href{https://cseweb.ucsd.edu/~mil070/jojo/}{Jojo}, for the inspiration to name this paper. This project is partly supported by NSF Award \#1750063.
\bibliography{custom}
\bibliographystyle{acl_natbib}

\appendix

\section{The BEIR Benchmark}
\label{sec:beir}

\begin{table*}[t!]
    \small
    \resizebox{\textwidth}{!}{\begin{tabular}{ l | l | l | c | c | c | c | c c c | c c }
        \toprule
         \multicolumn{1}{l}{\textbf{Split} ($\rightarrow$)} &
         \multicolumn{4}{c}{} &
         \multicolumn{1}{c}{\textbf{Train}}    &
         \multicolumn{1}{c}{\textbf{Dev}}    &
         \multicolumn{3}{c}{\textbf{Test}}   &
         \multicolumn{2}{c}{\textbf{Avg.~Word Lengths}} \\
         \cmidrule(lr){6-6}
         \cmidrule(lr){7-7}
         \cmidrule(lr){8-10}
         \cmidrule(lr){11-12}
           \textbf{Task ($\downarrow$)} &\textbf{Domain ($\downarrow$)} & \textbf{Dataset ($\downarrow$)} & \textbf{Title} & \textbf{Relevancy} & \textbf{\#Pairs} & \textbf{\#Query} & \textbf{\#Query} & \textbf{\#Corpus} & \textbf{Avg. D~/~Q } & \textbf{Query} & \textbf{Document} \\
         \midrule
    Passage-Retrieval    & Misc. & MS MARCO \shortcite{msmarco} & \xmark & Binary  & 532,761 &   ----  &   6,980   &   8,841,823      & 1.1 & 5.96  & 55.98  \\ \midrule[0.05pt] \midrule[0.05pt]
    Bio-Medical          & Bio-Medical & TREC-COVID \shortcite{roberts2020trec}  & \cmark & 3-level &   ----    &   ----  & 50     & 171,332   & 493.5& 10.60 & 160.77 \\
    Information          & Bio-Medical & NFCorpus \shortcite{boteva2016full}      & \cmark & 3-level & 110,575 &  324  & 323    & 3,633     & 38.2 & 3.30  & 232.26 \\
    Retrieval (IR)       & Bio-Medical & BioASQ \shortcite{tsatsaronis2015overview}       & \cmark & Binary  & 32,916 & ---- & 500    & 14,914,602& 4.7  & 8.05  & 202.61 \\ \midrule
    Question             & Wikipedia  & NQ  \shortcite{47761}           & \cmark & Binary  & 132,803  &   ----  & 3,452 & 2,681,468 & 1.2  & 9.16  & 78.88  \\
    Answering       & Wikipedia  & HotpotQA \shortcite{hotpotqa}     & \cmark & Binary  & 170,000 & 5,447 & 7,405  & 5,233,329 & 2.0  & 17.61 & 46.30  \\
     (QA)           &Finance& FiQA-2018 \shortcite{10.1145/3184558.3192301}  & \xmark & Binary  & 14,166  &  500  & 648    & 57,638    & 2.6  & 10.77 & 132.32 \\ \midrule
    Tweet-Retrieval      &Twitter& Signal-1M (RT)  \shortcite{signal}    & \xmark & 3-level &   ----    &   ----  & 97     & 2,866,316 & 19.6 & 9.30  & 13.93  \\ \midrule
    News      &News& TREC-NEWS  \shortcite{soboroff2019trec}    & \cmark & 5-level &   ----    &   ----  & 57     & 594,977 & 19.6 & 11.14  & 634.79  \\
    Retrieval      &News& Robust04 \shortcite{allan2004overview}    & \xmark & 3-level &   ----    &   ----  & 249   & 528,155 & 69.9 & 15.27  & 466.40  \\ \midrule
    Argument       & Misc. & ArguAna  \shortcite{wachsmuth:2018a}    & \cmark & Binary  &   ----    &   ----  & 1,406  & 8,674     & 1.0  & 192.98& 166.80 \\
    Retrieval   & Misc. & Touch\'e-2020 \shortcite{stein:2020v} & \cmark & 3-level &   ----    &   ----  & 49     & 382,545   & 19.0 & 6.55  & 292.37 \\ \midrule
    Duplicate-Question   &StackEx.& CQADupStack \shortcite{hoogeveen2015cqadupstack}  & \cmark & Binary  &   ----    &   ----  & 13,145 & 457,199   & 1.4  & 8.59  & 129.09 \\
    Retrieval            & Quora &  Quora        & \xmark & Binary  &   ----    & 5,000 & 10,000 & 522,931   & 1.6  & 9.53  & 11.44  \\ \midrule
    Entity-Retrieval     & Wikipedia  &  DBPedia  \shortcite{Hasibi:2017:DVT}     & \cmark & 3-level &   ----    &   67  & 400    & 4,635,922 & 38.2 & 5.39  & 49.68  \\ \midrule
    Citation-Prediction  & Scientific&  SCIDOCS  \shortcite{cohan-etal-2020-specter}     & \cmark & Binary  &   ----    &   ----  & 1,000  & 25,657    & 4.9  & 9.38  & 176.19 \\ \midrule
                         & Wikipedia  &  FEVER \shortcite{fever}       & \cmark & Binary  & 140,085 & 6,666 & 6,666  & 5,416,568 & 1.2  & 8.13  & 84.76  \\ 
    Fact Checking        & Wikipedia  & Climate-FEVER \shortcite{diggelmann2020climatefever} & \cmark & Binary  &   ----    &   ----  & 1,535  & 5,416,593 & 3.0  & 20.13 & 84.76  \\
                         & Scientific & SciFact  \shortcite{wadden-etal-2020-fact}     & \cmark & Binary  &   920      &   ----  &  300   & 5,183     & 1.1  & 12.37 & 213.63  \\
    \bottomrule
    \end{tabular}}
    \caption{Statistics of datasets in the BEIR benchmark. The table is taken from \citet{thakur2021beir}. Few datasets contain documents without titles. Relevancy indicates the query-document relation: binary (relevant, non-relevant) or graded into sub-levels. Avg.~D/Q indicates the average relevant documents per query. \vspace{-5mm}}
    \label{tab:dataset_stats}
\end{table*}

\paragraph{Datasets} We list the statistics of the BEIR benchmark in Table~\ref{tab:dataset_stats}. The 18 English  zero-shot evaluation datasets come from 9 heterogeneous retrieval tasks, including bio-medical information retrieval, question answering, tweet retrieval, news retrieval, argument retrieval, duplicate question retrieval, citation prediction, and fact checking. 

\paragraph{Metric} To measure effectiveness of search algorithms or retrieval models, the benchmark uses Normalized Discounted Cumulative Gain (nDCG) \cite{jarvelin2002cumulated} as the evaluation metric. We will give the definition of the metric in the following.

Given top $k$ retrieved documents $\{d_1,d_2,..,d_k\}$ with their relevance $\{r_1,r_2,..,r_k\}$ for a query, the traditional formula of discounted cumulative gain (DCG) accumulated at a particular rank position $k$ is defined in Equation \ref{eq:dcg}, where $r_i$ is 1 if $d_i$ is the ground truth otherwise 0.
\begin{equation}
\label{eq:dcg}
    DCG@K = \sum_{i=1}^K\frac{r_i}{log_2(i+1)}
\end{equation}

Since the length of ground truth list depends on the query, using DCG to compare the performance of retrieval models from one query to the next cannot be consistently achieved. Therefore, the discounted cumulative gain is normalized  (nDCG) as:
\begin{equation}
\label{eq:ndcg}
    nDCG@K = \frac{DCG@K}{IDCG@K}
\end{equation}
where IDCG@K is the DCG@K score for the list of relevant documents (ordered by their relevance) in the corpus up to position $k$. Since IDCG@K producs the maximum possible DCG through position $k$, the value of nDCG@K is in the range 0 to 1.

\section{Baselines}
\label{sec:baseline}
We use the baselines from the current BEIR leaderboard~\citep{thakur2021beir}. These baselines can be divided into four groups: dense retrieval, lexical retrieval, late interaction and re-ranking.

\paragraph{Dense Retrieval}
For dense retrieval, the baselines are the same dual-tower model as ours. We consider \textbf{DPR}~\cite{dpr}, \textbf{ANCE}~\cite{ance}, \textbf{TAS-B}~\cite{tasb} and \textbf{GenQ}~\cite{thakur2021beir} in this paper.
\begin{itemize}
    \item \textbf{DPR} uses a single BM25 retrieval example and in-batch examples as hard negative examples to train the model. Following \citet{thakur2021beir}, we use Multi-DPR as the baseline. The model is a BERT-base model and is trained on four QA datasets, including NQ~\cite{47761}, TriviaQA~\cite{triviaqa}, WebQuestions~\cite{berant-etal-2013-semantic} and CuratedTREC~\cite{baudivs2015modeling}.
    \item \textbf{ANCE}  constructs hard negative examples from an ANN index of the corpus. The hard negative training instances are updated in parallel during fine-tuning of the model. The model is a RoBERTa~\cite{liu2019roberta} model trained on MS-MARCO for 600k steps.
    \item \textbf{TAS-B} is trained with Balanced Topic Aware Sampling using dual supervision from a cross-encoder and a ColBERT model~\citep{colbert}. The model is trained with a combination of a pairwise Margin-MSE~\cite{hofstatter2021improving} loss and an in-batch negative loss function.
    \item \textbf{GenQ} fine-tunes a T5-base~\cite{t5} model on MS MARCO for 2 epochs and generate 5 queries for each document as additional training data to continue to fine-tune the TAS-B model. 
\end{itemize}

\paragraph{Lexical Retrieval}
Lexical retrieval is a score function for token matching calculated between two high-dimensional sparse vectors with token weights. BM25~\cite{bm25} is the most commonly used lexical retrieval function. We use the BM25 results reported in \citet{thakur2021beir} for comparison.

\paragraph{Late Interaction}
We also consider a late interaction baseline, namely \textbf{ColBERT}~\cite{colbert}. The model computes multiple contextualized embeddings for each token of queries and documents, and then uses a maximum similarity function to retrieve relevant documents. This type of matching requires significantly more disk space for indexes and has a higher latency.

\paragraph{Re-ranking}
Re-ranking based approaches use the output of a first-stage retrieval system (\eg BM25), and then re-rank the retrieved documents using a cross-encoder~\cite{nogueira2020passage}. In this paper, we use the \textbf{BM25+CE} baseline implemented in \citet{thakur2021beir} that uses BM25 to retrieve top-100 documents and a 6-layer MiniLM~\cite{minilm} model to further re-rank the recalled documents. 

\section{More Examples}
In addition to examples in Section~\ref{sec:case}, we provide more examples here, from a commonsense question answering dataset HotpotQA~\citep{hotpotqa}. %

\label{appendix_case}
\begin{figure}[htb]
    \centering
    \includegraphics[width=\linewidth]{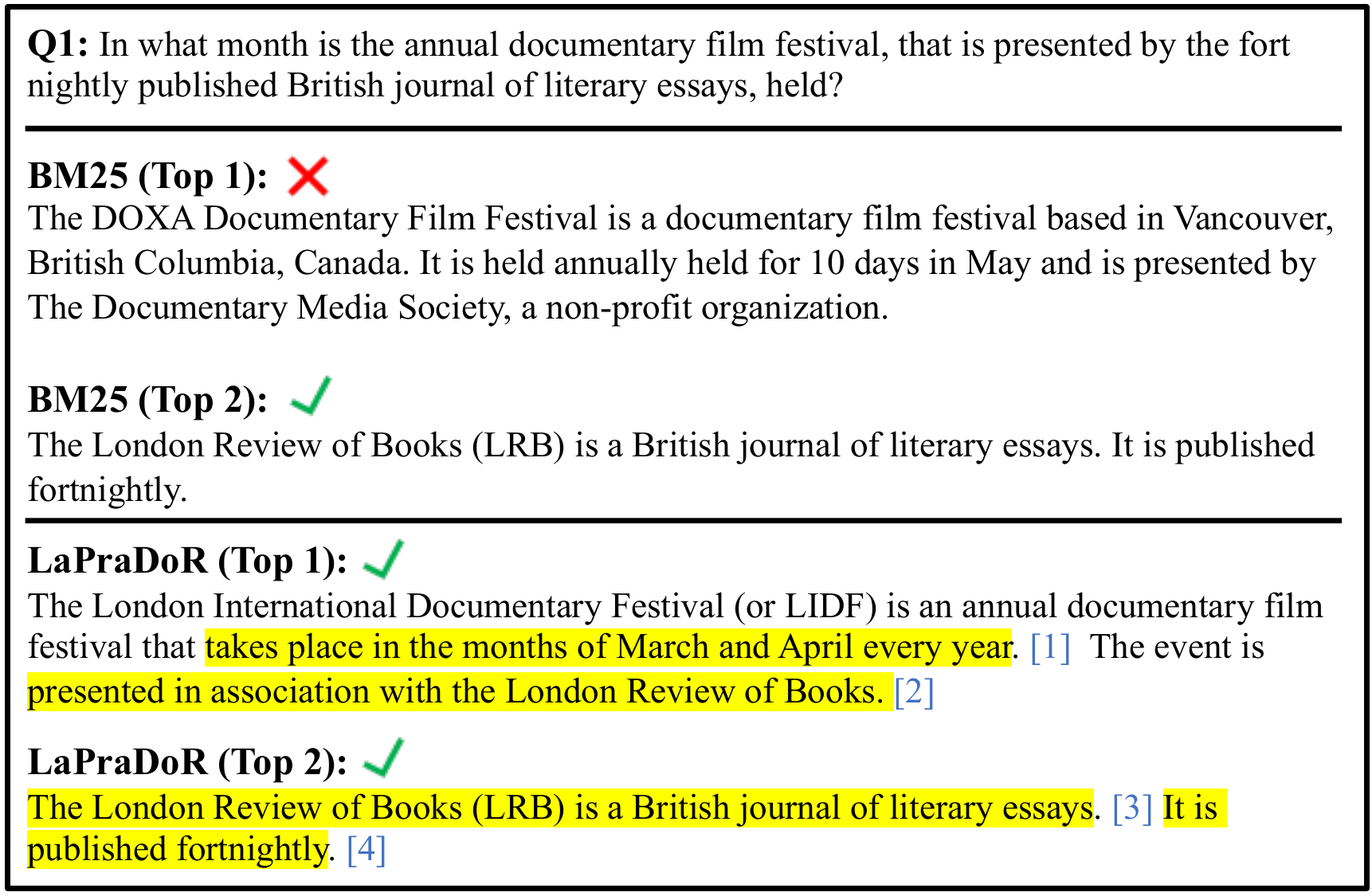}
    \includegraphics[width=\linewidth]{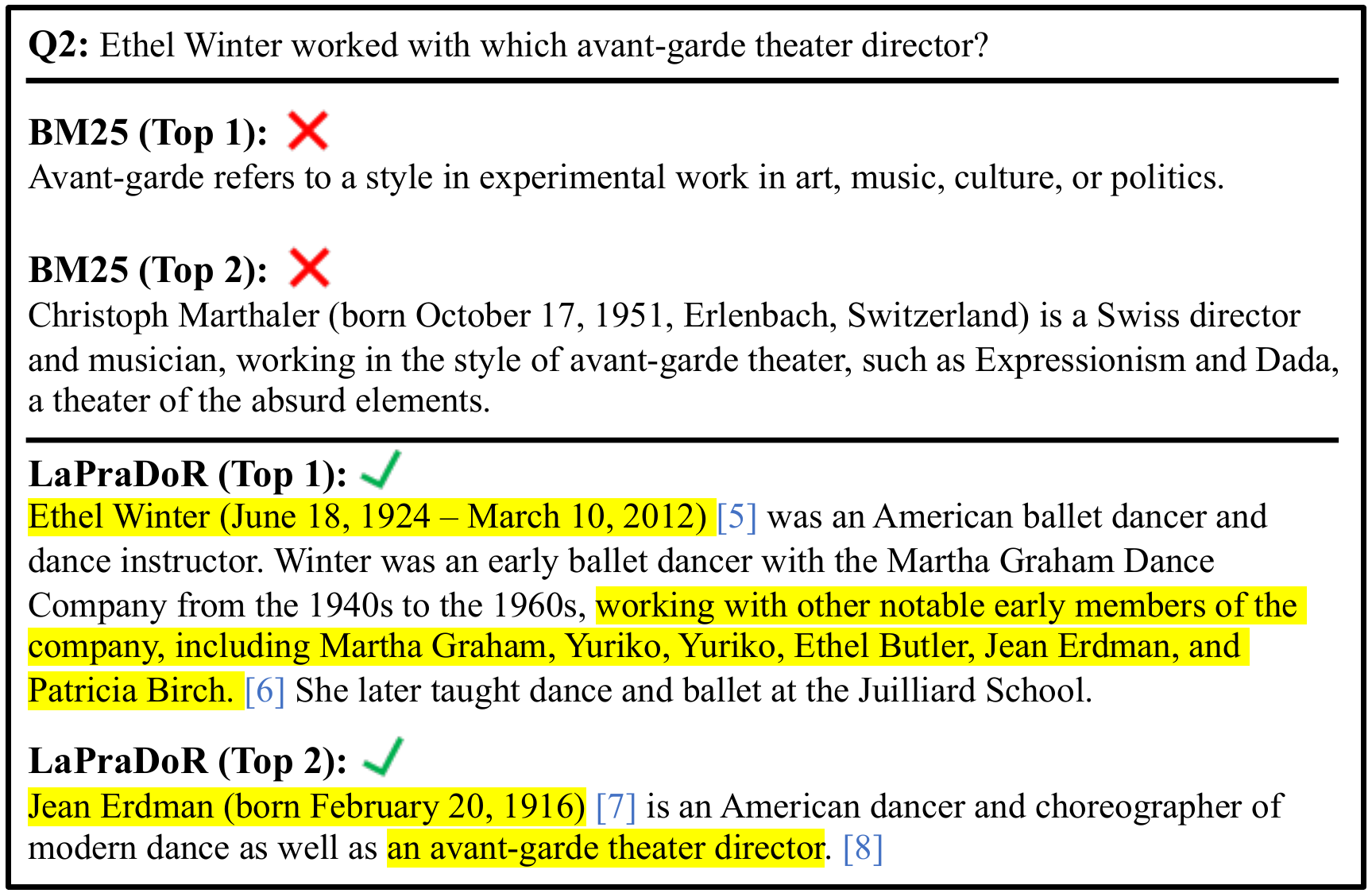}
    \caption{Examples from  the HotpotQA dataset~\citep{hotpotqa}. The key facts are highlighted. The reasoning path for Q1 is [3]$\rightarrow$[4]$\rightarrow$[2]$\rightarrow$[1] and for Q2 is [5]$\rightarrow$[6]$\rightarrow$[7]$\rightarrow$[8]. }
    \label{fig:case-hotpotqa}
\end{figure}

\end{document}